\documentclass[letterpaper, 10 pt, conference]{ieeeconf}  

\IEEEoverridecommandlockouts
\usepackage[T1]{fontenc} 
\author{Julian Ryde$^{1}$  and Xuchu (Dennis) Ding$^{2}$
\thanks{$^{1}$Julian Ryde is with United Technologies Research Center (UTRC). {\tt\small julian.ryde@utrc.utc.com} }
\thanks{$^{2}$Xuchu (Dennis) Ding is with Exyn Technologies Inc. {\tt\small xding@exyntechnologies.com}}
}%

\usepackage{dot2texi}
\usepackage{tikz}
\usetikzlibrary{shapes,arrows}

\usepackage[cmex10]{amsmath}
\usepackage{cite} 
\usepackage{fixltx2e}
\usepackage{multirow}

\usepackage{grffile} 
\graphicspath{{./images/}{./figures/}{./graphs/}}
\renewcommand{\baselinestretch}{0.97}

\begin{document}
\newlength{\imagewidth}
\setlength{\imagewidth}{0.5\textwidth}

\title{\LARGE \bf
RenderMap: Exploiting the Link Between Perception and Rendering for Dense Mapping
}

\maketitle 

\begin{abstract}
We introduce an approach for the real-time (2Hz) creation of a dense map and alignment of a moving robotic agent within that map by rendering using a Graphics Processing Unit (GPU).  This is done by recasting the scan alignment part of the dense mapping process as a rendering task.  Alignment errors are computed from rendering the scene, comparing with range data from the sensors, and minimized by an optimizer.  The proposed approach takes advantage of the advances in rendering techniques for computer graphics and GPU hardware to accelerate the algorithm. Moreover, it allows one to exploit information not used in classic dense mapping algorithms such as Iterative Closest Point (ICP) by rendering interfaces between the free space, occupied space and the unknown.  The proposed approach leverages directly the rendering capabilities of the GPU, in contrast to other GPU-based approaches that deploy the GPU as a general purpose parallel computation platform.  

We argue that the proposed concept is a general consequence of treating perception problems as inverse problems of rendering.  Many perception problems can be recast into a form where much of the computation is replaced by render operations.  This is not only efficient since rendering is fast, but also simpler to implement and will naturally benefit from future advancements in GPU speed and rendering techniques.  Furthermore, this general concept can go beyond addressing perception problems and can be used for other problem domains such as path planning.


\end{abstract}

\section{Introduction}


We explore using the native rendering pipeline of a GPU to solve perception problems, such as the generation of a dense map suitable for planning and localization of a moving agent within the dense map.  Instead of using the GPU as a general parallel computing platform, i.e., general purpose GPU (GPGPU) computing. 
We aim to exploit computer graphics rendering pipelines and take advantage of the vast amount of work already invested in improving this pipeline for simulation, video-gaming, movie, and many other industries. 

We argue that, rather than the utility of rendering being serendipitous, it is a product of a more fundamental premise: Perception is the inverse of rendering.  Perception is the act of constructing a model of the environment from sensory data whilst rendering is simulating the sensory data given a known environment. As such, many perception problems such as localization and mapping can be directly solved by repeatedly rendering the perceived environment and minimizing alignment errors within the rendering environment.

In order to use the computer graphics rendering pipeline directly to solve perception problems,  we recast perception problems to make them expressible as direct GPU rendering problems.
In particular, we focus on the real-time creation of high resolution textured polygon meshes from moving Lidar and RGB-D sensors, such as the Kinect or Asus Xtion Pro, and compute the alignment with respect to these meshes to localize a moving autonomous agent.
Such maps of the environment are vital for the navigation and control of mobile autonomous platforms, as well as useful for inspection tasks and monitoring.

The advantages of this approach are twofold: faster computation time and a reduced implementation burden.  The computer graphics industry exploits many sophisticated techniques to maximize utility of the rendering hardware to generate a scene.  Comparing to solving perception problems on a CPU or using GPGPU, an algorithm accelerated by rendering requires less code and will automatically benefit from improvements that will be made to the speed and accuracy of rendering techniques.  

Moreover, this approach allows us to take advantage of rendering techniques to exploit information either traditionally disregarded or difficult to utilize without specialized data structures.  As such the proposed algorithm exploits the complete information from sensors capable of outputting a point cloud such as a Lidar or a RGB-D sensor.
In the case of computing alignments for the pose of a moving agent within a dense map, we are able to utilize not only the information of "occupied space" such as point clouds, but also information of the implied "free space" between the occupied space and the unknown.  This is done by rendering interfaces between the free, occupied and unknown space as polygonal meshes and using them during alignments.  Note that computer graphics is particularly efficient and highly advanced for rendering polygonal meshes. We argue that comparing to traditional point-based approaches, such as the ICP algorithm \cite{besl1992}, the proposed algorithm is inherently capable of better accuracy and robustness since it utilizes information that is not available to ICP algorithms (i.e., not only the range data directly observed, but also range data derived from the known and observed free space).



The contributions of this paper are threefold.  First, we propose a localization and mapping pipeline that can rely on graphics rendering to efficiently generate a dense occupancy map and compute alignments within that map.  This pipeline is modular and both the alignment step and map-update step can be accelerated by rendering.  Second, we propose a render-accelerated alignment strategy via optimizing a carefully crafted cost function computed from rendering the scene and interfaces between free space, occupied space, and the unknown.  Third, we show that the concept of perception as the inverse of rendering is general, and can be applied to existing algorithms, such as ICP.  As such we propose a render accelerated ICP strategy for alignment.  

The focus of this paper is to present the process of recasting the scan alignment part of dense mapping process as rendering tasks, with steps required to set up the proper rendering environment such as meshification.  Each iteration of the scan alignment corresponds to a render of the scene in a candidate pose.  We use standard approaches for other aspects of the mapping process (i.e., voxelization, map-update and loop closure).  We believe that it is possible to recast the other parts of the process also as rendering problems, but will leave them as future work.

We note that there may be other robotics problems that can be cast as a rendering task and thus take advantage of rendering capabilities of the GPU.  Discussions on possible future applications of the proposed ideas to other problems can be found in Sec. \ref{sec_future_work}.

\section{Background and Related Work}
\label{sec:background}






Simultaneous Localization and Mapping (SLAM) has widely been used to solve the chicken-and-egg problem of localization and mapping in an unknown map. SLAM includes approaches based on Particle Filters \cite{sim2005vision}, extended Kalman Filters (EKF) \cite{huang2007convergence}, Probabilistic Hypothesis Density (PHD) \cite{mullane2011random}, and many others.  For a survey, see \cite{aulinas2008slam}.  

However, many approaches, such as the aforementioned ones, focus on producing localizations and maps with respect to a sparse set of features.  In this paper, we instead focus on 3D {\it dense map} (i.e., occupancy maps that represent the environment) construction from point-cloud data collected by a ranging sensor such as a RGB-D sensor or a Lidar, because we are interested in path planning and other tasks that can be performed on dense maps.  For dense map construction, ICP  \cite{besl1992} or Probabilistic methods \cite{kohlbrecher2014hector} have been commonly used for pose alignment between scans of point-cloud, and Bayesian occupancy-grid \cite{thrun2003learning} is a common approach to represent and update the map.
Some existing work exploits general purpose computing on the GPU for mapping smaller volumes, \cite{newcombe2013real,whelan2012kintinuous}, and more recently with OpenCL \cite{trifonov2013real} rather than directly using the GPU rendering functionality.
\cite{venugopal2013accelerating} uses the GPU for quick ray triangle intersections to process laser range data, \cite{olson2009real} tackles Lidar scan matching on the GPU, and \cite{peinecke2008lidar} exploits the fragment and vertex shaders. 

Texture mapping for robotics applications has received some attention \cite{rinnewitz2013automatic}, which mainly focuses on techniques to enhance the compression of textures.  The process of generating the mesh for texturing in \cite{wiemann2013automatic} is a modified marching cubes followed by mesh reduction and filtering and region growing of planes with texturing.

Random ball cover \cite{neumann2011real} is a data structure optimized for parallel processing for nearest neighbor search in dimensions 21-78.  It has proved successful for the low dimension 3D nearest neighbor search on multi-core systems.
When porting algorithms to the GPU there are numerous performance considerations. GPUs have many cores, but each core has only fast access to a small subset of limited memory, global memory access is comparatively slow, as such, communication between cores should be minimized.

The state-of-the-art for GPU mapping is perhaps \cite{neumann2011real} and others have achieved interactive frame rates \cite{engelhard2011real,fioraio2011realtime,henry2014rgb,huhle2008fly}. 
\cite{neumann2011real} note 1-2ms per ICP iteration and ICP convergence times of  20ms for scan-to-scan matching.
\cite{neumann2011real} comments that the nearest neighbor search dominates the run time so optimization should focus on this part of the procedure.


The closest approaches involving the direct comparison of the z-buffer are \cite{benjemaa1999fast,labsik2000depth}. Others \cite{fallon2012efficient} employ a low fidelity prior 3D model of the area of operation consisting of large planar sections and compare both color and depth for Monte Carlo based localization.
For considering the surface elements of a scene,  \cite{pfister2000surfels} implement surfels rather than a mesh to represent the map.




\section{RenderMap}
\label{sec:rendermap}

We propose an alignment and mapping framework based on rendering that addresses the following issues:
\begin{enumerate}
\item {\it Able to construct a dense map:} existing feature-based SLAM implementations as discussed in Sec. \ref{sec:background} would not be sufficient;
\item {\it Work with a large variety of operating environments:} existing methods, such as ICP would fail in environments with long edges such as an oblique wall; 
\item {\it Easily integrate between different sensor modalities:} difficult for existing methods such as ICP to fuse several different sensor data for alignment such as RGB and depth data;
\item {\it Efficient in processing speed:} bottleneck of ICP methods is the nearest neighbor search, which does not scale well with large point-cloud data; and
\item {\it Efficient in memory requirement:} bottleneck of Occupancy-grid methods for map-update is the size and resolution of the grid, which does not scale well with large environment.
\end{enumerate}



\begin{figure}
    \centering 
    \includegraphics[width=0.75\columnwidth]{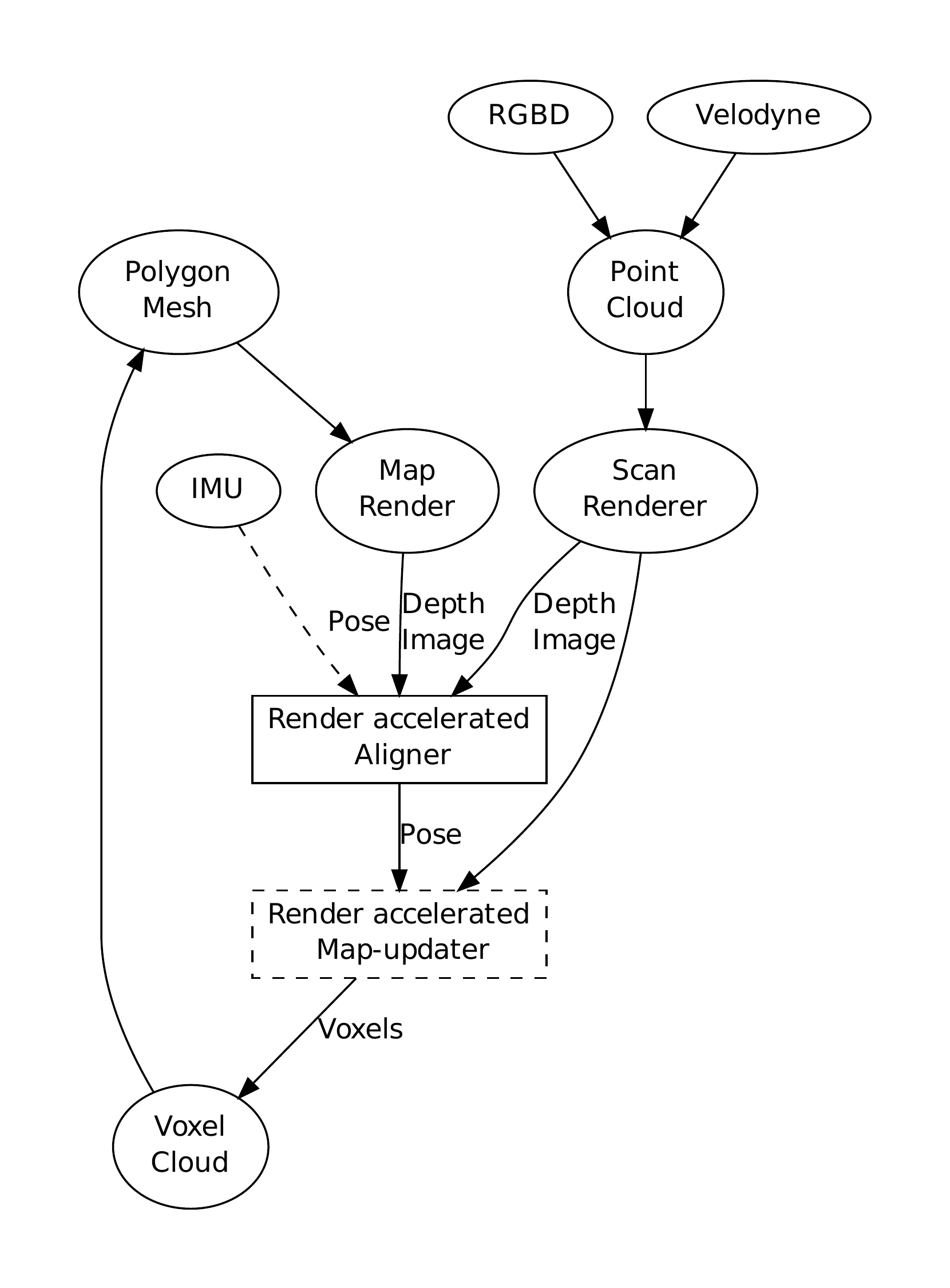}
    \caption{Overview of the full render-based dense-mapping system. Render accelerated alignment will be discussed in Sec. \ref{sec:opt-alignment} (optimization-based) and Sec. \ref{sec:render_icp} (ICP-based).  In this paper, we assume map-updater is standard, and will leave render-accelerated map-updater as future work (thus in shades).  IMU input is optional and can provide better initial estimates}
    \label{fig_render_SLAM_flow}
\end{figure}

The rest of the paper will discuss how issues (1)-(5) will be addressed in the proposed framework. An overview of the proposed framework is shown in Fig. \ref{fig_render_SLAM_flow}.  Note that the proposed framework is modular and both the map alignment module and the map update module can be render accelerated or not.  In this paper, we focus on two render accelerated scan-to-map alignment strategies.  The first is a direct numerical optimization on the per-pixel error metric between the z-buffers.  The second is an approximated version of point-to-plane ICP modified to benefit from GPU renders of the scene.   These are described in detail in the section \ref{sec:opt-alignment} and \ref{sec:render_icp} respectively. 
In this paper, we assume the use of traditional occupancy grids for map-updating.  Full render accelerated mapping pipeline, including render accelerated map-update which will be examined as future work.

\section{Optimization-based Z-buffer alignment in Rendered Environment}
\label{sec:opt-alignment}
There is strong motivation for direct optimization based alignment techniques \cite{blais1995registering} 
mainly because there are some fundamental limitations with ICP and other point based approaches.
For ICP, although very effective in certain situations, 
complications arise predominantly when establishing the association or correspondence among	 points.

Fast determination of the closest point is a major bottleneck in the processing speed of ICP. The closest point problem can be alleviated to some degree by sophisticated spatial data structures and proximity approximations (approximate or projective nearest neighbour).

As mentioned in Sec. \ref{sec:rendermap}, there are more fundamental problems with range alignment based on ICP which arise from its neglect of free space.  Rays emanating from a range sensor terminate with a distance measurement and the terminated points are fed into ICP alignment algorithms.  However, this is an approximation of the full ray and there is no information derived from rays as the fact that there is free space until the terminated point is useful information.  Moreover, if the rays do not return due to the lack of terminating points, they still contain valuable information as to the absence of obstacles that should included in the alignment process.  These are ignored in ICP algorithms.

\begin{figure}
    \centering 
    \includegraphics[width=0.45\columnwidth]{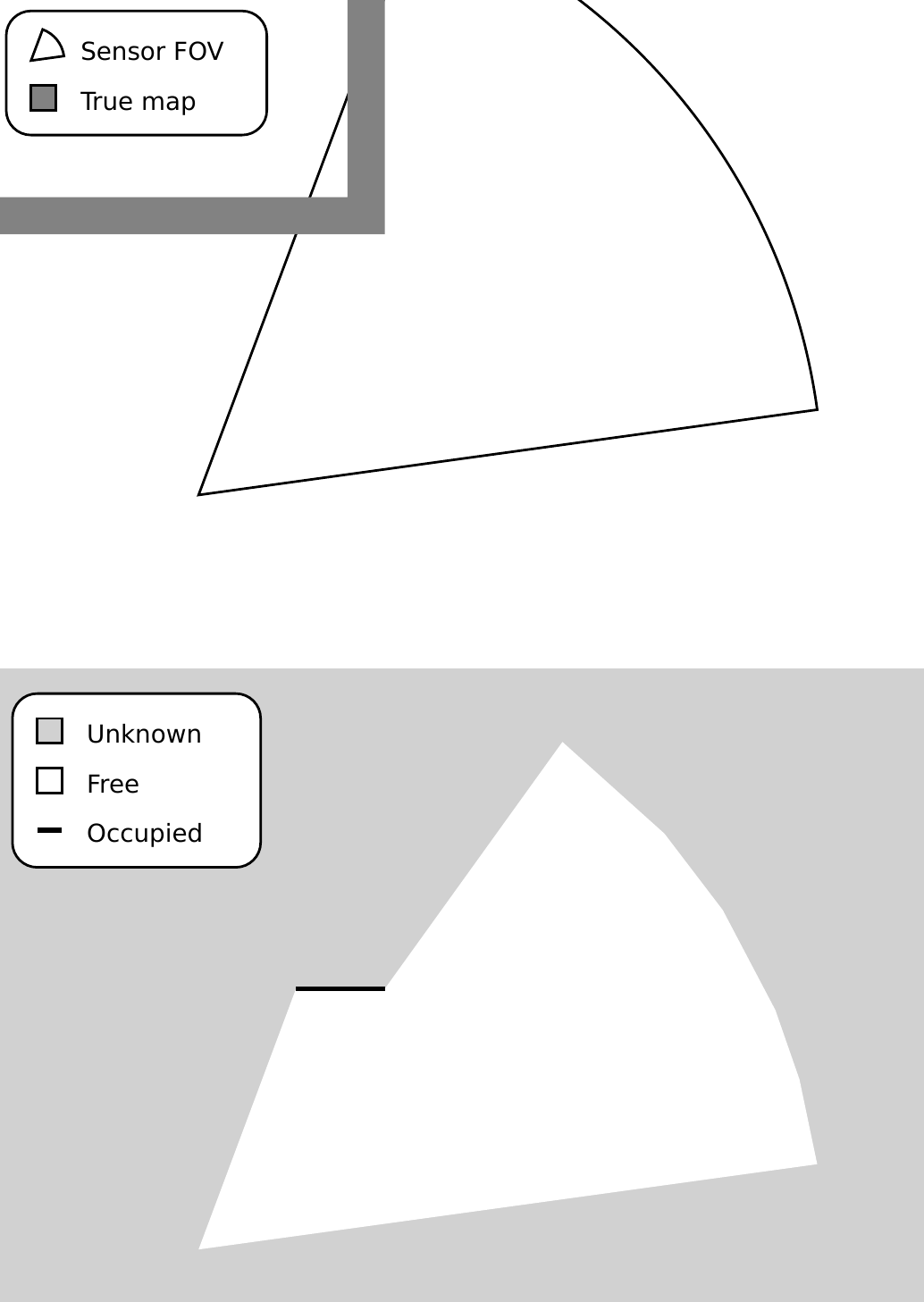}
    \includegraphics[width=0.45\columnwidth]{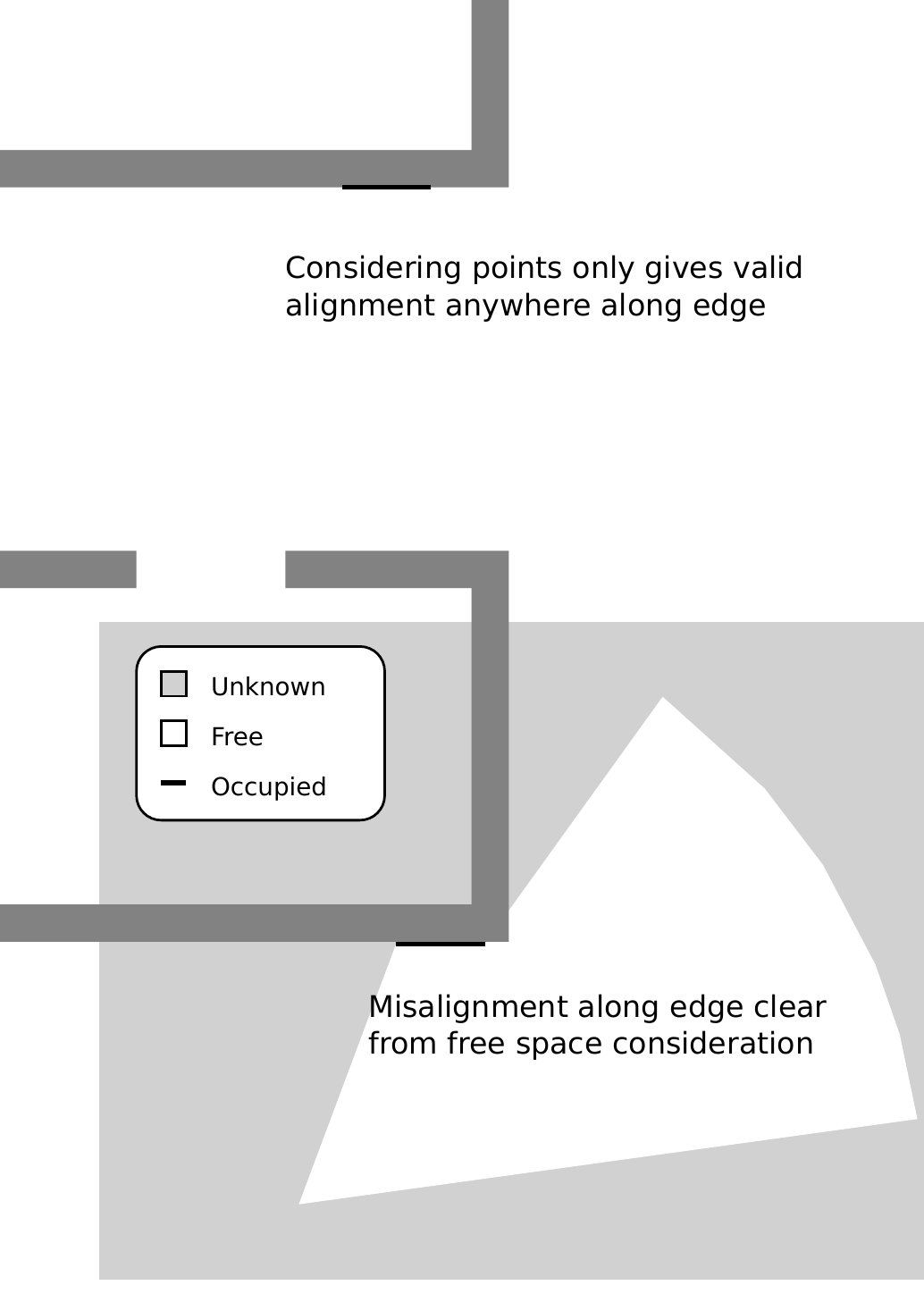}
    \caption{Diagram showing why methods that only consider points, such as ICP, fail to correctly align in certain situations.  Consider a sensor field-of-view as shown in the top-left figure.  Top-right figure shows that the alignment will have an error if only considering the points, as they can misalign anywhere along the edge of the wall.  Now consider aligning with all provided information including free space as shown in the bottom-left figure.  Bottom-right figure shows that misalignment will not occur in this case.}
    \label{fig_icp_failure_scenarios}
\end{figure}

This omission of information can cause problems in common scenarios such as observing the corner of the surface of a table or viewing the edge of building where only the ground plane and a obelisk wall and its edge are visible. As an example, consider the relatively common case of using RGBD sensors to match an observed corner of a table surface to a table (see Fig. \ref{fig_icp_failure_scenarios}).  Some practical examples of these types of failures will be shown in Sec. \ref{sec:results}. This may be addressed with an occupancy grid approach, but optimizing over a potential large span of the grid can be infeasible or not scalable and can be difficult to implement.

In this section we propose an alignment strategy based on using the Z-buffer obtained from each pixel after rendering the scene appropriately by the GPU.  
Z-buffer (also called depth-buffer) computation is an essential step of the GPU rendering pipeline to determine what to render for each pixel, and Z-buffer images can be obtained as the output of the rendering process.  The Z-buffer image is similar to a depth image where each pixel indicates the distance between the camera and the rendered object.  

Alignment between scan-to-scan or scan-to-map can be seen as the problem of minimizing the Z-buffer misalignment error.    As such, we carefully craft a cost function that can be evaluated in a rendered environment.  Instead of using a grid-type approach, we simply render the partitions between occupied, free and unknown space as polygonal mesh surfaces.  Therefore, this approach can take into account all observed information when aligning to the map.  The challenge of this approach is to design the cost function carefully to consider all cases where each pixel may land on free, occupied or unknown space.  

Although we focus on range-only alignment in results shown this paper, another advantage of direct render based optimization is the ease with which color information can be incorporated into the cost function and hence factored into the alignment optimization.  

\subsection{Meshification and Rendering}
\begin{figure}
    \centering 
    \includegraphics[width=0.45\columnwidth]{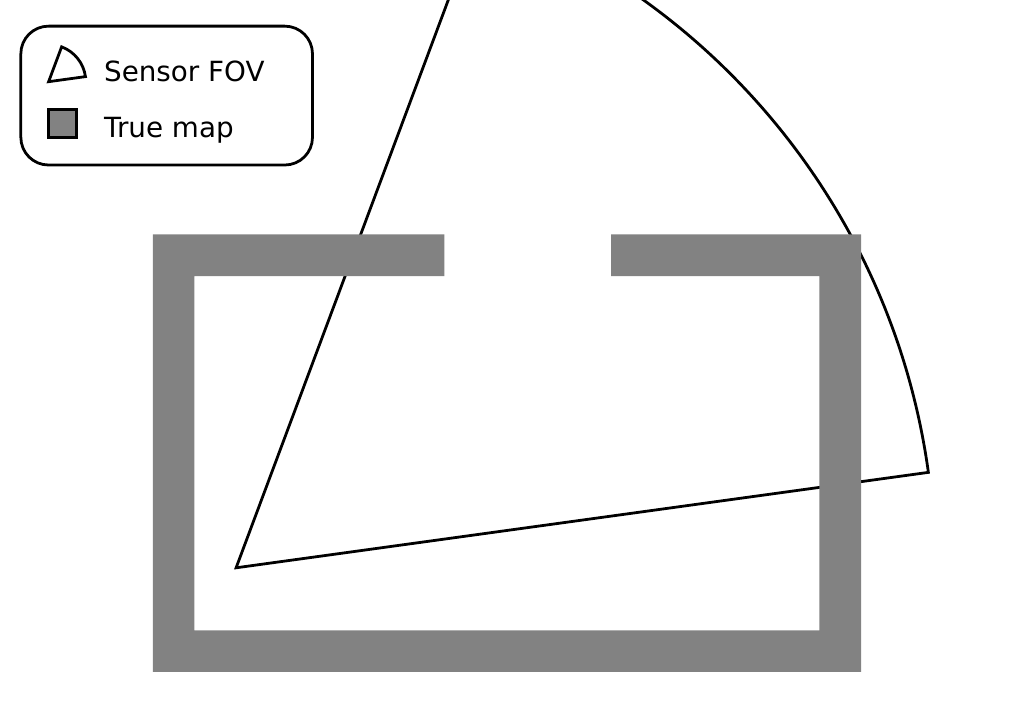}
    \includegraphics[width=0.5\columnwidth]{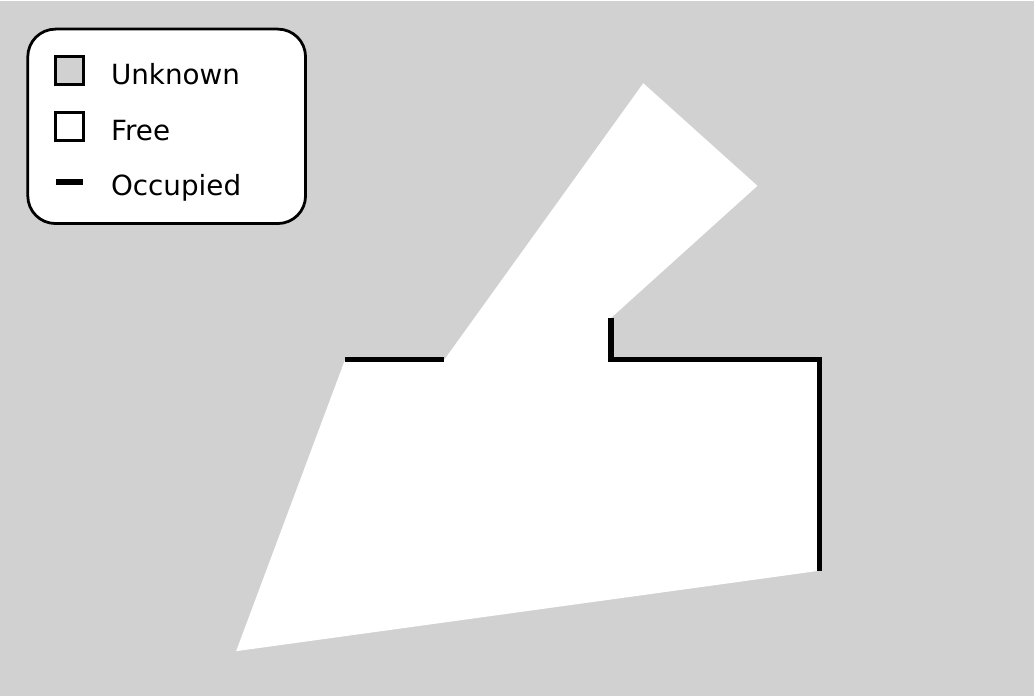}
    \caption{2D example illustrating the free, occupied and unknown space and interfaces between these volumes for a particular scan of a range sensor.}
    \label{fig_space_partition}
\end{figure}

The space in the environment can be partitioned into three volumes: free, unknown, and occupied space.  Between these three volumes there are three partitions or interfaces: free-unknown, free-occupied,  and unknown-occupied.  An example of such volumes and partitions is shown in Fig. \ref{fig_space_partition}. Converting the volumetric representation of the space into its partitioning interfaces enables representation of the volumes on the graphics card, which is particularly designed to render surfaces (textured polygon meshes) quickly.  By rendering a depth image and comparing with sensor range image we can analyze each pixel as a measurement ray emanating from within the sensor field-of-view towards the scene, and reward or penalize based on if the ray landed inside the free space, the occupied space or unknown space.

\begin{figure}
    \centering 
    \includegraphics[width=0.5\columnwidth]{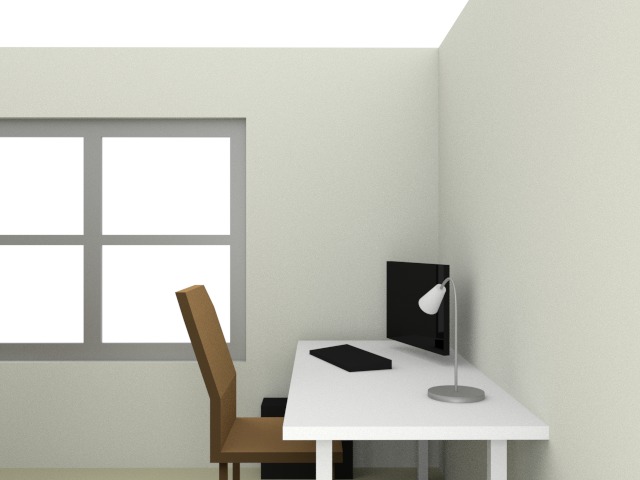}%
    \includegraphics[width=0.5\columnwidth]{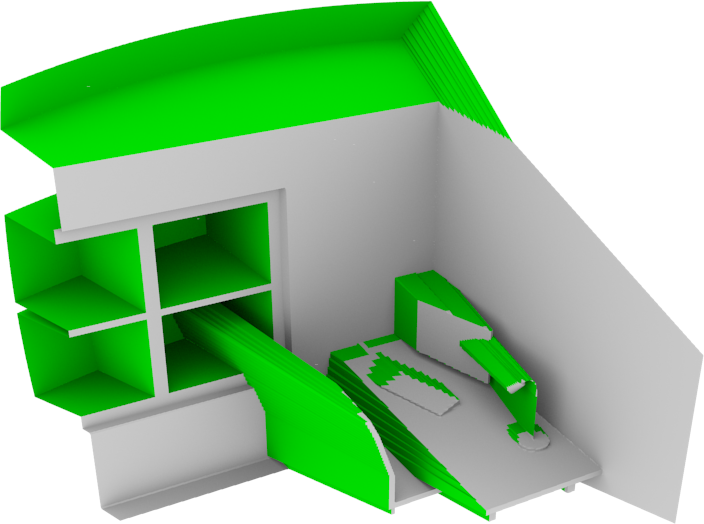}
    \caption{Render of polygonal mesh space-partitioning interfaces for a single range image in a synthetic scene.  The ceiling has been removed from the scene to aid visualization.  Green surfaces represent the interface between free space and unknown space.  Gray surfaces represent the interface between free and occupied space.  White is the background color, indicating area with no space information data. }
    \label{fig_meshification_render}
\end{figure}

\begin{figure}
    \centering 
    \includegraphics[width=0.47\columnwidth]{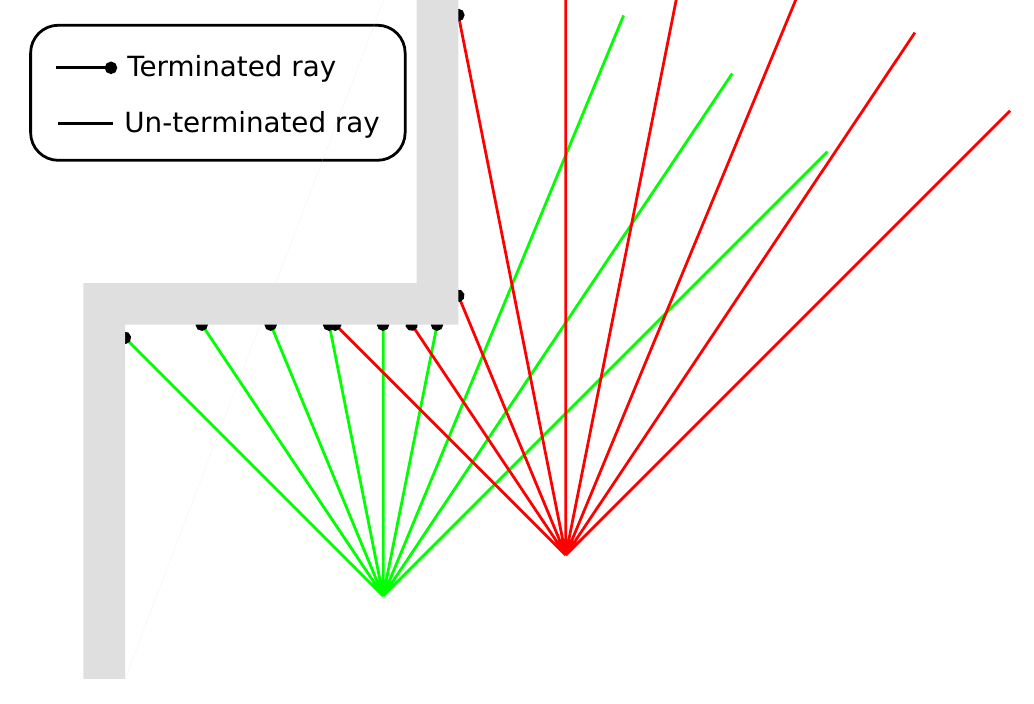}
    \includegraphics[width=0.47\columnwidth]{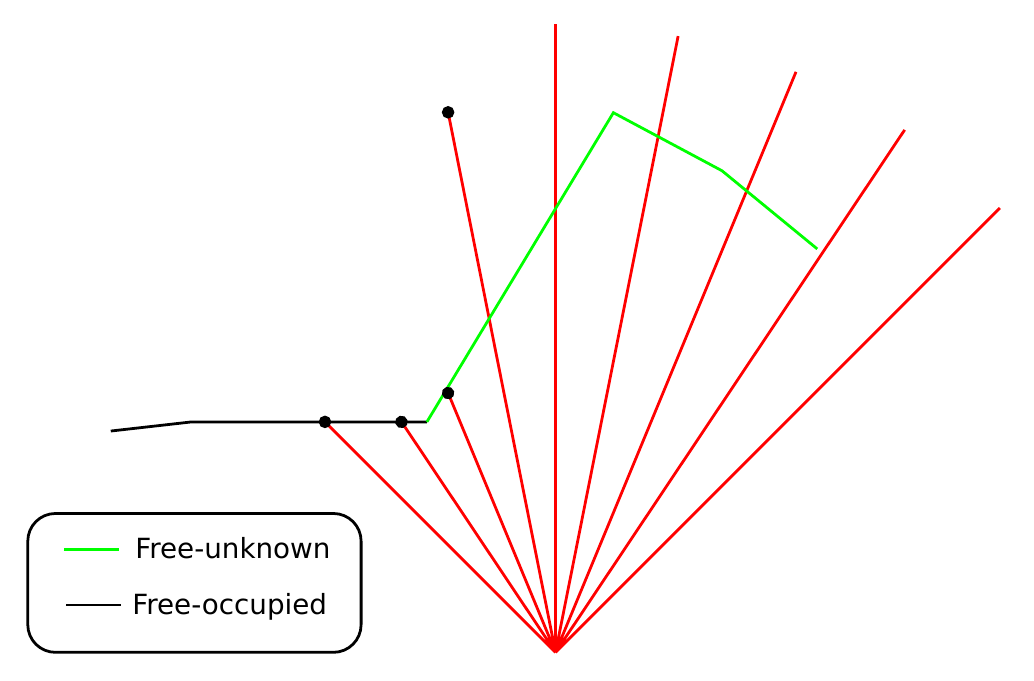}    
    \caption{2D diagram of meshification of surfaces.  Left: range measurement rays from two
     consecutive scans (left is first scan and right is the second).  Right: 2D surfaces created from the first scan, which will be used for alignment to compute the pose of the second scan.  In this case, the free-unknown interface created by the first scan is "seen" by the second scan.  This can be useful for alignment - if the alignment is such that the free-unknown interface is not seen, then there is an error as it does not agree with the space information given from the first scan.}
    \label{fig_meshification_diagram}
\end{figure}

An example of the mesh generated from a single range image is rendered in Fig. \ref{fig_meshification_render}.  Note that the interface between occupied and unknown space are usually "behind the view-port", therefore hidden and not needed to be rendered.  The mesh is generated from a range image or other structured range data by generating quadrilaterals between adjacent 2 by 2 pixels in the range image.  Pixels for which there was no range readings are assumed to indicate free space up to a sensor dependent maximum. For example, maximal range of $4$ meters is used for structured light RGB-D sensors such as the Mcrosoft Kinect and ASUS Xtion Pro.  No-return pixel quadrilaterals are drawn at this maximal distance and colored green.  Quadrilaterals that are too large (> 0.1m) and span range discontinuities are also colored green to denote the free-unknown interface.  The remaining quadrilaterals approximate the observed free-occupied interface and are colored gray.  Space with no information (i.e. space beyond all the surfaces) are colored with a background color (we used white).  This process in 2D is summarized in Fig. \ref{fig_meshification_diagram}.

From Fig. \ref{fig_meshification_diagram}, we can see how the full space information taken from each scan can be used for alignment.  Alignment error can be determined by rendering the scan to be aligned at a candidate pose, e.g., if the space is supposed to free but now observed from Z-buffer value of the pixel that it is occupied or unknown, then there is an alignment error.


\subsection{Cost Function}
\label{sec_cost_function}

\begin{table*}[t]
  \caption{Table for computing the error $E(X)$, as determined by the depth
      value of each pixel $d(p_s)$ in current scan $Z_s$ and the depth value
      $d(p_r)$ of the corresponding pixel in the map or the previous scan
      $Z_r$, summed over all pixels. $\Delta z=d(p_r) - d(p_s)$ and $\epsilon = 0.1m$ is the inlier-outlier rejection threshold .
      The N/A in the table indicate situations that do not arise.  The "ignore"
      cells in the table indicates situations where there is insufficient
      information to determine a reward or a cost.
}
  \centering
  \begin{tabular}{cc|c|c|c|c|}
\cline{3-6}
& & \multicolumn{4}{ c| }{Pixel $p_s$ of current Scan $Z_s$} \\ \cline{3-6}
& & Free-occupied & Free-unknown & Unknown-occupied & $\mathtt{NaN}$ \\ \cline{1-6}
\multicolumn{1}{ |c  }{\multirow{4}{*}{
\begin{tabular}[x]{@{}c@{}}Pixel $p_r$ of\\Map/Previous\\Scan $Z_r$\end{tabular}
} } &
\multicolumn{1}{ |c| }{Free-occupied} 
&\begin{tabular}[x]{@{}c@{}} Penalize (+1) if $|\Delta z| > \epsilon$ 
\\Reward (-1) if $|\Delta z|  \leq \epsilon$ 
\end{tabular}
& \begin{tabular}[x]{@{}c@{}}Penalize (+1) if $\Delta z \leq 0 $\\ Reward (-1) if $\Delta z > 0$\end{tabular}
& N/A & ignore \\ \cline{2-6}
\multicolumn{1}{ |c  }{}                        &
\multicolumn{1}{ |c| }{Free-unknown} 
& \begin{tabular}[x]{@{}c@{}}Penalize (+1) if $\Delta z > 0$\\ Reward (-1) if $\Delta z \leq 0$\end{tabular}
& ignore & N/A & ignore \\ \cline{2-6}
\multicolumn{1}{ |c  }{}                        &
\multicolumn{1}{ |c| }{Unknown-occupied} & ignore & ignore & N/A & ignore \\ \cline{2-6}
\multicolumn{1}{ |c  }{}                        &
\multicolumn{1}{ |c| }{$\mathtt{NaN}$} & ignore & ignore & N/A & ignore \\ \cline{1-6}
\end{tabular}
  \label{tab:pixel_classification_logic}
\end{table*}

After rendering the scene with all the interfaces between free, occupied and unknown space, we aim to design an objective function to have both a large region of convergence and a unique extremum at or very close to the true pose. 
Due to the separations of the range measurements, 
corresponding rays from two scans will hit slightly different points.  This issue is mostly addressed by the meshification rendering process as discussed in the previous sub-section, since we are rendering surfaces instead of points.  But due to errors from the sensor and the fact that the range measurements may not be sufficiently dense, the rendered surfaces will always be an approximation of the actual environment. Thus, the objective function would not be guaranteed to have an extremum exactly at the true pose and our goal is to design the cost function to have the unique extremum as close as possible to the true pose.


Consider the problem of computing the relative pose from the current scan to the previous scan or the map. Denoting the pose as $X=(x,y,z, \theta_x, \theta_y, \theta_z)$, we propose the error function 
\begin{equation}
    E(X) = C (R(T(X)), Z_s),
\end{equation}
where $R(Y)$ is the Z-buffer image of the GPU render of the previous scan at a particular camera pose $Y$, $T$ transforms the pose $X$ of the observer to the pose $Y$ of the camera, the Z-buffer image of the meshes rendered from the current scan is denoted as $Z_s$, and $C(Z_1, Z_2)$ is a function that compares two Z-buffer images at each pixel in the image and produces a scalar output.   Note that this error function is defined the same way for scan-to-scan alignments and scan-to-map alignments (in this case, consider $R(Y)$ as output of rendering of the map at camera pose $Y$).

The key is to design the function $C$ to be a good metric for alignment error.  Denote $Z_r = R(T(X))$ as the Z-buffer of the map rendered at a candidate pose $X$, note that each pixel of $Z_r$ can either land at free-occupied surface, free-unknown surface or unknown-occupied surface and the depth value corresponds to the distance to the surface, or the value can be a $\mathtt{NaN}$ (not a number) which indicates that the ray emanating from the pixel does not hit any rendered surfaces (i.e., it reached an area with no space information).  Similarly, each pixel of the $Z_s$ can be either on the free-occupied or free-unknown surface, created from the points in the current scan.  By definition, this Z-buffer image will not contain any pixels on the unknown-occupied space (as they cannot be seen).  It also should not contain any $\mathtt{NaN}$ unless there are dead pixels in the depth image.

The complete logic to determine the cost function is summarized in Table \ref{tab:pixel_classification_logic}. This table is created from desired behaviors of a correct alignment.  For example, if a pixel in the map/previous scan lands on the free-unknown surface and the pixel in the current scan lands on the free-occupied, then $\Delta z > 0 $ indicates an alignment error, since there is now occupied space in area which was cleared previously.  

\begin{figure}
    \centering 
    \includegraphics[width=0.32\columnwidth]{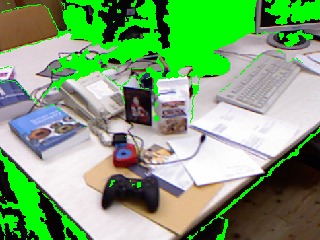}
    \includegraphics[width=0.32\columnwidth]{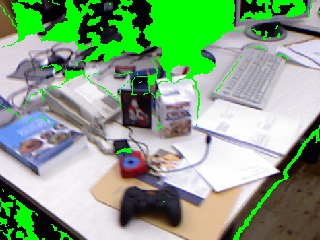}
    \includegraphics[width=0.32\columnwidth]{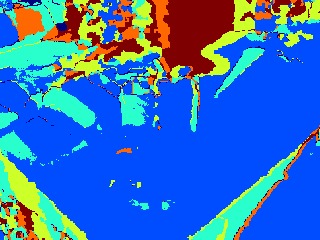}
    \caption{Example images showing pixel classifications colored based on Table \ref{tab:pixel_classification_logic}.  Green in the top images represents the free-unknown interface. Top-left: rendering of the current scan.  Top-right: rendering of the previous scan at the initial pose.  Bottom: each pixel is labeled with color for their comparison classification.  
    Cyan: penalized outliers on free-occupied interface and when $|\Delta z| > \epsilon$;
    Blue: rewarded inliers on free-occupied interface; 
    Yellow and Orange: unterminated rays in either current scan or map render;
    Red: both are on the free-unknown interface.
}
    \label{fig_pixel_classification_logic}
\end{figure}

\begin{figure}
    \centering 
    \includegraphics[width=1\columnwidth]{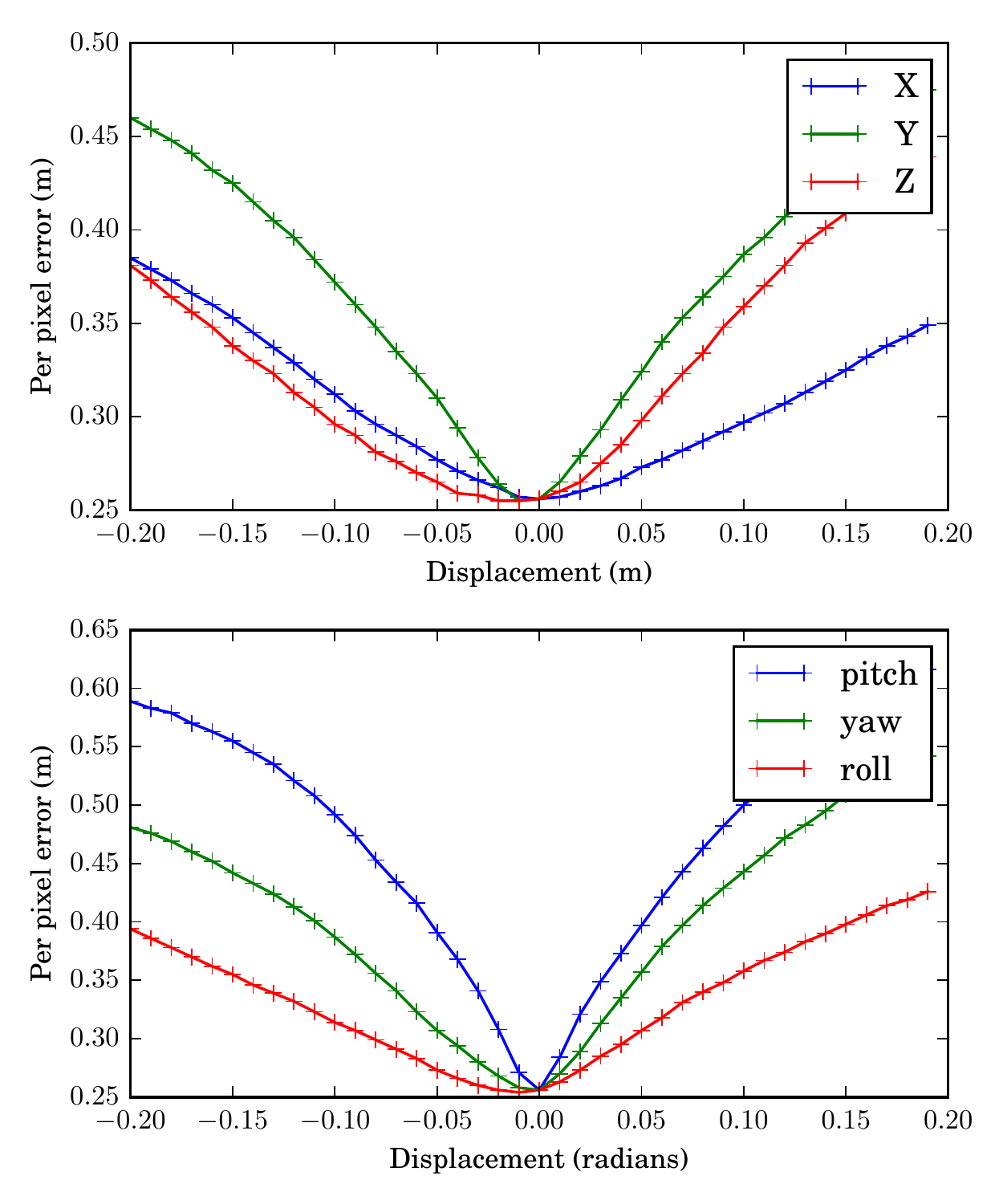}
    \caption{Cost function $E(X)$ plotted near the true pose (red dot) for the scan as shown in Fig. \ref{fig_pixel_classification_logic}. Top figure is positional displacement in meters and bottom figure is angular displacement in radians.  
    }
    \label{fig_alignment_minima}
\end{figure}

The algorithm to compute $E(X)$ is as follows.  Every ray from the sensor either terminates on a surface and is a line segment or continues indefinitely and is a ray.  We first determine the classification of the pixel (which surface does it terminates on?), then the per-pixel-error error value is derived based on Table \ref{tab:pixel_classification_logic}.  The total error $E(X)$ is computed by summing the per-pixel error for all pixels. Fig. \ref{fig_pixel_classification_logic} shows the process running on an RGB-D dataset (\emph{fr3/long\_office\_household} dataset \cite{sturm12iros} at scan 400).   We also plot the cost function around the true pose at the same scan in Fig.~\ref{fig_alignment_minima}.  

\subsection{Optimization}

The objective function is vital for effective optimization. Ideally, the objective function should be smooth, have low noise, and have a large region of convergence with a unique minimum near the true-pose.  However, similarly to ICP and other optimization-based approaches, the issue of multiple local-minima is generally possible and should be addressed. The cost function as discussed in the previous section is designed so that the region of convergence is sufficiently large.  There is at least one local minimum outside of the region of convergence, in the case where the camera pose is pointing away from the map (in this case, all pixel in $Z_r$ are $\mathtt{NaN}$s).  In the relevant pose-space of interest (i.e., about 20-30cm around the true pose) and running the data-set as described in Sec. \ref{sec:results}, we have indeed seen the cost function to exhibit a unique global minimum very close to the true pose (as shown in Fig. \ref{fig_alignment_minima}) in all scans.  However, this is only supported by empirical evidence and the cost function may exhibit multiple local minima for some scenes.

We realize that there are opportunities (and more work to be done) to adjust the shape and characteristics of the objective function to improve convergence for optimization and reducing non-ideal local minima, such as adjusting the per-pixel-error calculation to exclude large changes in value when changing the pose, post processing the depth images by Gaussian blurring to reduce noise in the cost function, and introducing costs and rewards based on colors of corresponding pixels (similar to \cite{pascoe2015farlap}).  Furthermore, adjusting the weight for penalties and rewards within the cost function for different classifications in Table \ref{tab:pixel_classification_logic} can help make the descent sharper and remove the local minimum when the camera is positioned away from the map.

For the cost function presented in this paper, we use a number of derivative-free optimization methods such as Nelder-Mead and Powell's method \cite{powell1973search} aided by initial rapid gradient-descent.  Note that as shown in Fig. \ref{fig_alignment_minima}, the cost drop-off along some coordinate axes are more rapid than others.  This indicates a Rosenbrock-like function \cite{rosenbrock1960automatic} and as such can be efficiently solved by Coordinate Descent or Adaptive Coordinate Descent approaches when near the "valley".  In order to minimize the number of evaluations of the cost function (each requires one render), we use a hybrid optimizer that combines a number of the above approaches. The results in this paper are obtained by using an optimizer with initial gradient-descent, then a number of Nelder-Mead iterations plus some Coordinate Descent iterations once near the minimum.





\section{Render Accelerated ICP}
\label{sec:render_icp}
We now describe using rendering techniques to speed up existing algorithms, such as ICP.
For point-to-plane ICP, a set of corresponding points and their associated normals are needed.  To achieve this we recast the ICP algorithm into a render problem.  The most time consuming aspect of ICP is the point-to-point correspondence.  We relax the nearest point requirement and instead use an approximation of nearest point along the observer ray, sometimes referred to as projective correspondence \cite{rusinkiewicz2001efficient}.  
This distance is usually a good approximation to the nearest neighbor, improving as the surface normal aligns with the view direction.

This approximated ICP problem can be solved by considering the difference between the Z-buffer rendered from a camera pose and the depth map of the scan.  
Given these observed ray correspondences the errors are minimized to produce the next pose in the iteration.   This process is a follows:
\begin{enumerate}
    \item Acquire points $X_0$ from laser scan/RGB-D image;
    \item Convert $X_0$ to polygon mesh to create a map $M$;
    \item Convert $X_1$ to a depth map $Z_1$;
    \item Render $M$ from initial guess pose $P_0$ to get depth map $Z_M$ of
        same size as $Z_1$;
    \item Compare $Z_M$ and $Z_1$ to generate a cost $C = \sum_{ij} |Z_M - Z_1|$; and
    \item Adjust $P_1$ to minimize $C$ via ICP given the pixel-to-pixel
        correspondences in the depth map/z-buffer.
\end{enumerate}

Given two sets of corresponded points we can solve for the rotation and translation.  According to \cite{rusinkiewicz2001efficient} the matching strategy has the largest effect on the convergence and hence speed of ICP.  This projection-based matching results in a slightly worse performance per iteration but is faster to compute versus closest-point (especially for GPU renders), and it requires the point-to-plane error metric \cite{chen1992object}. 

Every pixel in the depth map can be converted to its position in space.  Only those with a z-difference of less than a tolerance are considered associated and passed to the ICP minimization.  The values from the Z-buffer $z_b$  are converted to real perpendicular distances from the camera plane as
\(Z = z_0 / ( 1 - z_b (1 - z_0/z_\infty))\) 
with $z_0$ and $z_\infty$ being the near and far clip distances.  The pixel coordinates for each range pixel in the render buffer are then converted to their real world position.


Experimentally, it was determined that point-to-point ICP does not work well with the projective association.  This becomes apparent when the ICP point-to-point projection based correspondence lines for a typical alignment are shown.  Point-to-plane ICP works well with the projective correspondence.  Fig. \ref{fig_normals_image} outlines the process for generating the normals necessary for point-to-plane ICP and how they are rendered to accelerate the normal lookup.



\subsection{Z-buffer ICP Alignment}
In general for the purposes of robot navigation, point-to-plane ICP is better than point-to-point ICP.  This is mainly because the distribution of Lidar points is not uniform over the surrounding surfaces.  Oblique surfaces create an anisotropic distribution of Lidar returns.
Inspecting the Lidar returns from the ground plane reveals rings of returns due to the interaction of the scan pattern and the oblique ground plane.
The rings of returns present significant problems to point-to-point ICP because they are preferentially aligned.  Since these rings are artifacts of the scan pattern rather than features of the environment, their alignment does not contribute to accurate pose.
Point-to-plane ICP is not affected by this because points are aligned to the ground plane rather than the ring points.

\begin{figure*} 
    \centering 
    \includegraphics[width=0.32\textwidth]{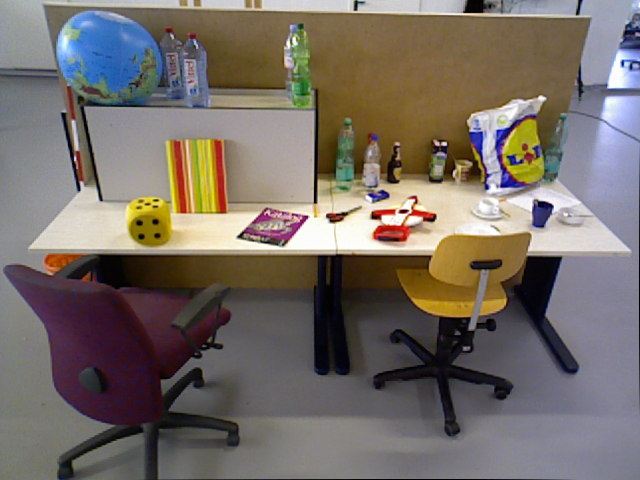}
    \includegraphics[width=0.32\textwidth]{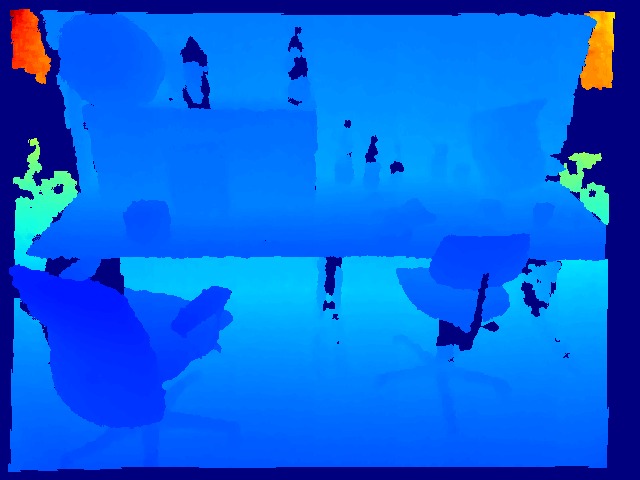}
    \includegraphics[width=0.32\textwidth]{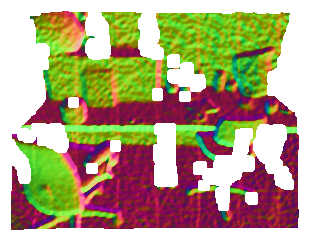}
    \caption{First and second, the RGB image and depth image from an RGB-D sensor.  Third, the corresponding surface normal image in which the surface normals are encoded by pixel RGB value.  Depth image surface normals are estimated by (\ref{eq_depth_image_normal}).  The normal map enables very fast point-to-plane ICP because point-to-plane ICP requires pairs of corresponding points and their normals.  Coloring by normal means that associated points and their normal can be quickly looked up by rendering from the estimated sensor point of view.}
    \label{fig_normals_image}
    \label{fig_depth_normal}
\end{figure*}

Point-to-plane ICP requires a fast method for calculating the surface normals directly from a depth image.  The method is similar to shape from shading \cite{horn1989obtaining} and more recently to \cite{holz2012real}.
The depth image is smoothed with a Gaussian and then the gradient computed by central differences across and down the depth image $z = f(x, y)$ to give the partial derivatives $\partial z/\partial x$ and $\partial z/ \partial y$.  These are converted into the surface normal $N(x, y)$ via the cross product,
\begin{equation}
N(x, y) = 
\begin{bmatrix} 0 \\ \partial y \\ \partial z \end{bmatrix}
\times
\begin{bmatrix} \partial x \\ 0 \\ \partial z \end{bmatrix}
=
\begin{bmatrix} 
      \partial y \, \partial z \\
      \partial z \, \partial x \\
    - \partial y \, \partial x \\
\end{bmatrix}.
    \label{eq_depth_image_normal}
\end{equation}

Fig.~\ref{fig_depth_normal} shows an example of surface normals estimated from a single depth image.

\subsection{Sequential Depth Image Alignment}
%
We now outline the render based sequential depth image alignment algorithm.
Consider a series of range images $D_i$ taken sufficiently quickly that the pose displacement between each is small. Small means that the pose difference should be within the convergence region of ICP which is typically less than 1m and 20 degrees.
This algorithm calculates the pose difference between two sequential depth images $D_i$ and $D_{i+1}$.
\begin{enumerate}
    \item Downsample $D_i$ by factor of 2 and filter out points that are too far;
    \item Calculate normals $N_i$ of $D_i$ via (\ref{eq_depth_image_normal});
    \item Render the points of $D_i$ colored by their normals.
    This associates depth image points with those in the rendered view and their surface normal;
    \item Perform point to plane ICP of points $D_{i+1}$ to calculate the
        relative pose transform matrix between $D_i$ and $D_{i+1}$; and
    \item Move render camera to new pose and repeat from 3.
\end{enumerate}
This is repeated to return the pose differences between all sequential $D_i$.
With any odometry type method, long term drift is a problem.  This error can be reduced by  matching until the pose displacement is significant rather than matching strictly sequentially.

\section{Results}
\label{sec:results}
Experiments are performed on sequential alignments of successive point clouds using RenderMap alignment (Sec. \ref{sec:opt-alignment}).  Although less accurate than aligning each scan to the entire map, this so called scan-to-scan matching allows for a more controlled comparison between ICP and RenderMap.  By only considering each scan to a previous scan, the effects of the map update mechanism and cumulative errors are avoided and a direct comparison of the localization efficacy is possible.

ICP is perhaps the most popular scan matching algorithm and has numerous variants.  As such, we benchmark for accuracy against a standard point-to-point ICP algorithm with outlier rejection threshold of $0.2m$.


%



For optimum performance, rendering is performed off screen and all anti-aliasing is disabled.  
The render window is 240 by 320 pixels.
The GPU for these experiments was an Nvidia Geforce GTX 765M with render speed of around 300 million points per second. 
Rendering a scene of a single 240 $\times$ 320 RGB-D camera is typically 1ms, this rate might fall for large scenes but this can be managed by the near and far culling planes of the view frustum. 


Experiments are conducted on the publicly available Freiburg datasets \cite{sturm12iros} based on RGB-D sensor measurements.  We ran sequential scan matching on one of the more representative and challenging datasets \emph{fr3/long\_office\_household}.  Our optimizer based scan matching achieved an average error of 0.02m per second of data which is similar to the errors of contemporary scan matching methods listed in \cite{kerl2013dense}.  It is worth noting that alignment optimizer is running on the depth data alone.  On the same dataset the median error drift rate for our alignment optimizer was 0.009m/s versus 0.019m/s for ICP.

\begin{figure}
    \centering 
    \includegraphics[width=1\columnwidth]{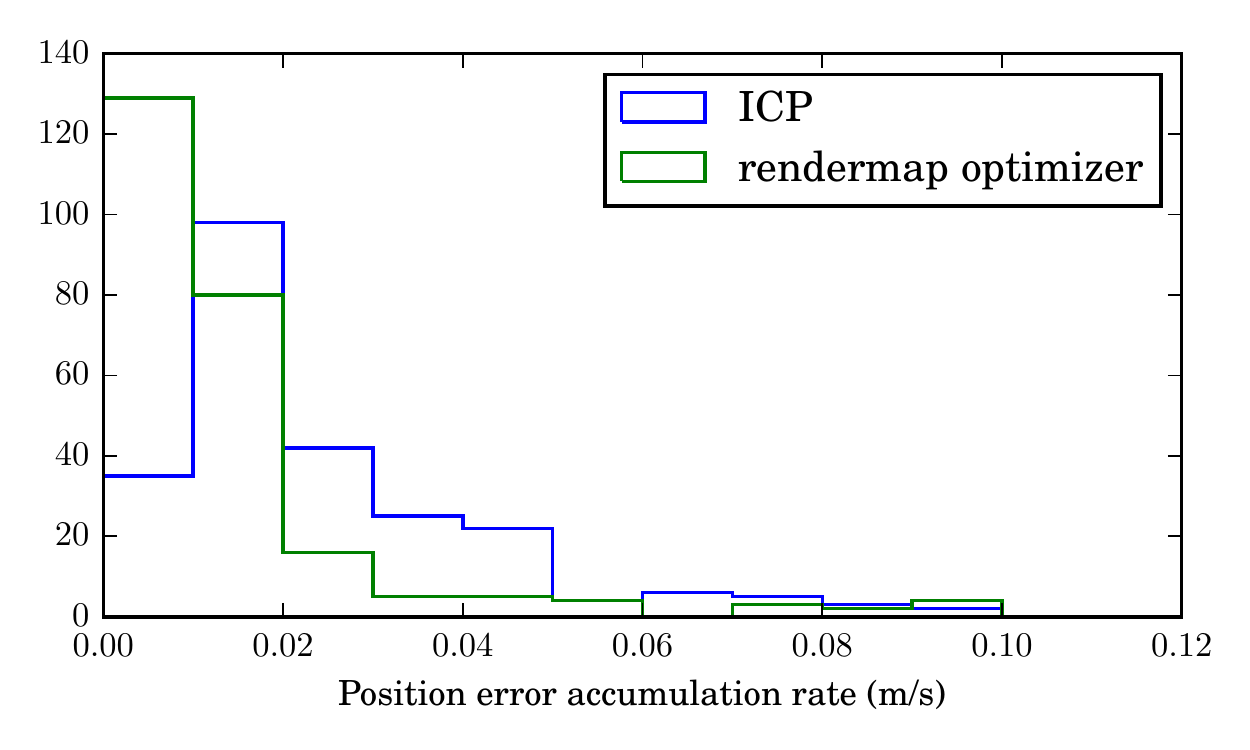}
    \caption{Comparison of the distribution of error drift rates for scan matching all data in the  \emph{fr3/long\_office\_household} dataset versus that for ICP.  In RenderMap, about 85\% of the time the position error is less than 1cm whereas for ICP it is only about 28\%.}
    \label{fig_pose_error_distribution}
\end{figure}

The result is shown in Fig. \ref{fig_pose_error_distribution}.  Note that many scans in this dataset failed to align with ICP but aligned with RenderMap approach.  
The average number of evaluations of the cost function (each require one rendering) across all sequential matches is about 700.  The average displacement between scans across all sequential matches is about 0.15m.   The time required to match each scan would depend on the graphics card.  For one such as Nvidia Geforce GTX 765M the render time is well below 1ms per frame, thus our approach is shown to be capable of running in real time since we are matching scans one second apart.


%
\section{Conclusions and Future Work}
\label{sec_future_work}
We present a number of improvements to dense 3D mapping.  One is changing from representing three volumes for dense mapping, (occupied, free and unknown) to one that represents the interfaces between these volume via conversion to a polygon mesh (meshification).  Once represented in this manner, alignment and map updates can be accomplished via GPU renders and therefore greatly accelerated.  This is different to general purpose GPU (GPGPU) approach which takes conventional algorithms and leverages parallel processing to speed up their computation.  Second, we devise a cost function that incorporates, in addition to the standard free-known interfaces the free-space information, the free-unknown boundaries and information from pixels that have no range reading.
Rather than point based methods such as ICP we perform optimization on this cost function to find the pose of the best alignment for each incoming scan by performing many repeated GPU renders. We tested proposed algorithms on a public range image dataset (Freiburg \cite{sturm12iros}) for sequential scan-to-scan matching.  These experiments demonstrated superior accuracy and robustness when compared to ICP.  Robustness is particularly important for many robotics applications where localization reliability is paramount for reliable continuous operation.  We plan to test the algorithm with additional public datasets in the future to better benchmark the algorithm with other dense mapping algorithms.

This work lays the pathway for future work on large scale 3D dense mapping via efficient scan-map matching as well as the potential for fast render accelerated map updates. We note that there are many more robotics problems (not limited to perception) that stand to benefit from GPU rendering, including point-to-plane ICP, ray traced map updates, object detection, nearest neighbor search, path planning, and collision checking.  Thus, the concept proposed in this paper may be general and can have large impact on a variety of algorithms used to solve autonomy problems.



\bibliographystyle{IEEEtran}
\bibliography{../bib/RenderPerception}

\begin{thebibliography}{10}
\providecommand{\url}[1]{#1}
\csname url@samestyle\endcsname
\providecommand{\newblock}{\relax}
\providecommand{\bibinfo}[2]{#2}
\providecommand{\BIBentrySTDinterwordspacing}{\spaceskip=0pt\relax}
\providecommand{\BIBentryALTinterwordstretchfactor}{4}
\providecommand{\BIBentryALTinterwordspacing}{\spaceskip=\fontdimen2\font plus
\BIBentryALTinterwordstretchfactor\fontdimen3\font minus
  \fontdimen4\font\relax}
\providecommand{\BIBforeignlanguage}[2]{{%
\expandafter\ifx\csname l@#1\endcsname\relax
\typeout{** WARNING: IEEEtran.bst: No hyphenation pattern has been}%
\typeout{** loaded for the language `#1'. Using the pattern for}%
\typeout{** the default language instead.}%
\else
\language=\csname l@#1\endcsname
\fi
#2}}
\providecommand{\BIBdecl}{\relax}
\BIBdecl

\bibitem{besl1992}
P.~Besl and N.~McKay, ``A method for registration of 3-{D} shapes,'' \emph{IEEE
  Transactions on Pattern Analysis and Machine Intelligence}, vol.~14, no.~2,
  pp. 239--256, 1992.

\bibitem{sim2005vision}
R.~Sim, P.~Elinas, M.~Griffin, J.~J. Little \emph{et~al.}, ``Vision-based slam
  using the rao-blackwellised particle filter,'' in \emph{IJCAI Workshop on
  Reasoning with Uncertainty in Robotics}, vol.~14, no.~1, 2005, pp. 9--16.

\bibitem{huang2007convergence}
S.~Huang and G.~Dissanayake, ``Convergence and consistency analysis for
  extended kalman filter based slam,'' \emph{Robotics, IEEE Transactions on},
  vol.~23, no.~5, pp. 1036--1049, 2007.

\bibitem{mullane2011random}
J.~Mullane, B.-N. Vo, M.~D. Adams, and B.-T. Vo, ``A random-finite-set approach
  to bayesian slam,'' \emph{Robotics, IEEE Transactions on}, vol.~27, no.~2,
  pp. 268--282, 2011.

\bibitem{aulinas2008slam}
J.~Aulinas, Y.~R. Petillot, J.~Salvi, and X.~Llad{\'o}, ``The slam problem: a
  survey.'' in \emph{CCIA}.\hskip 1em plus 0.5em minus 0.4em\relax Citeseer,
  2008, pp. 363--371.

\bibitem{kohlbrecher2014hector}
S.~Kohlbrecher, J.~Meyer, T.~Graber, K.~Petersen, U.~Klingauf, and O.~von
  Stryk, ``Hector open source modules for autonomous mapping and navigation
  with rescue robots,'' in \emph{RoboCup 2013: Robot World Cup XVII}.\hskip 1em
  plus 0.5em minus 0.4em\relax Springer, 2014, pp. 624--631.

\bibitem{thrun2003learning}
S.~Thrun, ``Learning occupancy grid maps with forward sensor models,''
  \emph{Autonomous robots}, vol.~15, no.~2, pp. 111--127, 2003.

\bibitem{newcombe2013real}
\BIBentryALTinterwordspacing
R.~NEWCOMBE, S.~Izadi, D.~Molyneaux, O.~Hilliges, D.~Kim, J.~Shotton, P.~Kohli,
  A.~Fitzgibbon, S.~Hodges, and D.~Butler, ``Real-time camera tracking using
  depth maps,'' September~19 2013, {US} Patent App. 13/775,165. [Online].
  Available: \url{http://www.google.com/patents/US20130244782}
\BIBentrySTDinterwordspacing

\bibitem{whelan2012kintinuous}
T.~Whelan, M.~Kaess, M.~Fallon, H.~Johannsson, J.~Leonard, and J.~McDonald,
  ``Kintinuous: Spatially extended kinectfusion,'' 2012.

\bibitem{trifonov2013real}
D.~Trifonov, ``Real-time high resolution fusion of depth maps on {GPU},''
  \emph{arXiv preprint arXiv:1311.7194}, 2013.

\bibitem{venugopal2013accelerating}
V.~Venugopal and S.~Kannan, ``Accelerating real-time {LiDAR} data processing
  using {GPUs},'' in \emph{Circuits and Systems (MWSCAS), 2013 IEEE 56th
  International Midwest Symposium on}.\hskip 1em plus 0.5em minus 0.4em\relax
  IEEE, 2013, pp. 1168--1171.

\bibitem{olson2009real}
E.~B. Olson, ``Real-time correlative scan matching,'' in \emph{Robotics and
  Automation, 2009. ICRA'09. IEEE International Conference on}.\hskip 1em plus
  0.5em minus 0.4em\relax IEEE, 2009, pp. 4387--4393.

\bibitem{peinecke2008lidar}
N.~Peinecke, T.~Lueken, and B.~R. Korn, ``{LiDAR} simulation using graphics
  hardware acceleration,'' in \emph{Digital Avionics Systems Conference, 2008.
  DASC 2008. IEEE/AIAA 27th}.\hskip 1em plus 0.5em minus 0.4em\relax IEEE,
  2008, pp. 4--D.

\bibitem{rinnewitz2013automatic}
K.~O. Rinnewitz, T.~Wiemann, K.~Lingemann, and J.~Hertzberg, ``Automatic
  creation and application of texture patterns to {3D} polygon maps,'' in
  \emph{Intelligent Robots and Systems (IROS), 2013 IEEE/RSJ International
  Conference on}.\hskip 1em plus 0.5em minus 0.4em\relax IEEE, 2013, pp.
  3691--3696.

\bibitem{wiemann2013automatic}
T.~Wiemann, K.~Lingemann, and J.~Hertzberg, ``Automatic map creation for
  environment modelling in robotic simulators,'' \emph{Proc. ECMS}, 2013.

\bibitem{neumann2011real}
D.~Neumann, F.~Lugauer, S.~Bauer, J.~Wasza, and J.~Hornegger, ``Real-time
  {RGB-D} mapping and {3-D} modeling on the {GPU} using the random ball cover
  data structure,'' in \emph{Computer Vision Workshops (ICCV Workshops), 2011
  IEEE International Conference on}.\hskip 1em plus 0.5em minus 0.4em\relax
  IEEE, 2011, pp. 1161--1167.

\bibitem{engelhard2011real}
N.~Engelhard, F.~Endres, J.~Hess, J.~Sturm, and W.~Burgard, ``Real-time {3D}
  visual slam with a hand-held {RGB-D} camera,'' in \emph{Proc. of the {RGB-D}
  Workshop on {3D} Perception in Robotics at the European Robotics Forum,
  Vasteras, Sweden}, vol. 2011, 2011.

\bibitem{fioraio2011realtime}
N.~Fioraio and K.~Konolige, ``Realtime visual and point cloud slam,'' in
  \emph{Proc. of the {RGB-D} workshop on advanced reasoning with depth cameras
  at robotics: Science and Systems Conf.(RSS)}, vol.~27, 2011.

\bibitem{henry2014rgb}
P.~Henry, M.~Krainin, E.~Herbst, X.~Ren, and D.~Fox, ``{RGB-D} mapping: Using
  depth cameras for dense {3D} modeling of indoor environments,'' in
  \emph{Experimental Robotics}.\hskip 1em plus 0.5em minus 0.4em\relax
  Springer, 2014, pp. 477--491.

\bibitem{huhle2008fly}
B.~Huhle, P.~Jenke, and W.~Stra{\ss}er, ``On-the-fly scene acquisition with a
  handy multi-sensor system,'' \emph{International Journal of Intelligent
  Systems Technologies and Applications}, vol.~5, no.~3, pp. 255--263, 2008.

\bibitem{benjemaa1999fast}
R.~Benjemaa and F.~Schmitt, ``Fast global registration of {3D} sampled surfaces
  using a multi-z-buffer technique,'' \emph{Image and Vision Computing},
  vol.~17, no.~2, pp. 113--123, 1999.

\bibitem{labsik2000depth}
U.~Labsik, R.~Sturm, and G.~Greiner, ``Depth buffer based registration of
  free-form surfaces.'' in \emph{VMV}.\hskip 1em plus 0.5em minus 0.4em\relax
  Citeseer, 2000, pp. 121--128.

\bibitem{fallon2012efficient}
M.~F. Fallon, H.~Johannsson, and J.~J. Leonard, ``Efficient scene simulation
  for robust monte carlo localization using an rgb-d camera,'' in
  \emph{Robotics and Automation (ICRA), 2012 IEEE International Conference
  on}.\hskip 1em plus 0.5em minus 0.4em\relax IEEE, 2012, pp. 1663--1670.

\bibitem{pfister2000surfels}
H.~Pfister, M.~Zwicker, J.~Van~Baar, and M.~Gross, ``Surfels: Surface elements
  as rendering primitives,'' in \emph{Proceedings of the 27th annual conference
  on Computer graphics and interactive techniques}.\hskip 1em plus 0.5em minus
  0.4em\relax ACM Press/Addison-Wesley Publishing Co., 2000, pp. 335--342.

\bibitem{blais1995registering}
G.~Blais and M.~D. Levine, ``Registering multiview range data to create {3D}
  computer objects,'' \emph{Pattern Analysis and Machine Intelligence, IEEE
  Transactions on}, vol.~17, no.~8, pp. 820--824, 1995.

\bibitem{sturm12iros}
J.~Sturm, N.~Engelhard, F.~Endres, W.~Burgard, and D.~Cremers, ``A benchmark
  for the evaluation of rgb-d slam systems,'' in \emph{Proc. of the
  International Conference on Intelligent Robot Systems (IROS)}, Oct. 2012.

\bibitem{pascoe2015farlap}
G.~Pascoe, W.~Maddern, A.~D. Stewart, and P.~Newman, ``Farlap: Fast robust
  localisation using appearance priors,'' in \emph{Robotics and Automation
  (ICRA), 2015 IEEE International Conference on}.\hskip 1em plus 0.5em minus
  0.4em\relax IEEE, 2015, pp. 6366--6373.

\bibitem{powell1973search}
M.~J. Powell, ``On search directions for minimization algorithms,''
  \emph{Mathematical Programming}, vol.~4, no.~1, pp. 193--201, 1973.

\bibitem{rosenbrock1960automatic}
H.~Rosenbrock, ``An automatic method for finding the greatest or least value of
  a function,'' \emph{The Computer Journal}, vol.~3, no.~3, pp. 175--184, 1960.

\bibitem{rusinkiewicz2001efficient}
S.~Rusinkiewicz and M.~Levoy, ``Efficient variants of the {ICP} algorithm,'' in
  \emph{3-D Digital Imaging and Modeling, 2001. Proceedings. Third
  International Conference on}.\hskip 1em plus 0.5em minus 0.4em\relax IEEE,
  2001, pp. 145--152.

\bibitem{chen1992object}
Y.~Chen and G.~Medioni, ``Object modelling by registration of multiple range
  images,'' \emph{Image and vision computing}, vol.~10, no.~3, pp. 145--155,
  1992.

\bibitem{horn1989obtaining}
B.~K. Horn, \emph{Obtaining shape from shading information}.\hskip 1em plus
  0.5em minus 0.4em\relax MIT press, 1989.

\bibitem{holz2012real}
D.~Holz, S.~Holzer, R.~B. Rusu, and S.~Behnke, ``Real-time plane segmentation
  using {RGB-D} cameras,'' in \emph{RoboCup 2011: Robot Soccer World Cup
  XV}.\hskip 1em plus 0.5em minus 0.4em\relax Springer, 2012, pp. 306--317.

\bibitem{kerl2013dense}
C.~Kerl, J.~Sturm, and D.~Cremers, ``Dense visual slam for rgb-d cameras,'' in
  \emph{Intelligent Robots and Systems (IROS), 2013 IEEE/RSJ International
  Conference on}.\hskip 1em plus 0.5em minus 0.4em\relax IEEE, 2013, pp.
  2100--2106.

\end{thebibliography}

\end{document}